\documentclass[10pt,twocolumn,letterpaper]{article}

\usepackage[pagenumbers]{cvpr} % To force page numbers, e.g. for an arXiv version

\usepackage{graphicx}
\usepackage{amsmath}
\usepackage{amssymb}
\usepackage{booktabs}

\usepackage{color}
\usepackage{multirow}
\usepackage{booktabs}
\usepackage{algorithm,algpseudocode}
\usepackage{setspace}
\usepackage{caption}
\usepackage{subcaption}
\usepackage{listings}
\usepackage{adjustbox}
\usepackage{bbm}
\usepackage{verbatim}
\usepackage{courier}
\usepackage[pagebackref,breaklinks,colorlinks]{hyperref}

\usepackage[capitalize]{cleveref}
\crefname{section}{Sec.}{Secs.}
\Crefname{section}{Section}{Sections}
\Crefname{table}{Table}{Tables}
\crefname{table}{Tab.}{Tabs.}

\def\cvprPaperID{1893} % *** Enter the CVPR Paper ID here
\def\confName{CVPR}
\def\confYear{2023}

\begin{document}

%%%%%%%%% TITLE - PLEASE UPDATE
\title{SeqCo-DETR: Sequence Consistency Training for \\ Self-Supervised Object Detection with Transformers}

\author{
Guoqiang Jin$^{1}$, Fan Yang$^{2,3}$, Mingshan Sun$^{1}$, Ruyi Zhao$^{1}$, Yakun Liu$^{1}$,\\
Wei Li$^{1}$, Tianpeng Bao$^{1}$, Liwei Wu$^{1}$, Xingyu Zeng$^{1}$,  Rui Zhao$^{1}$ \\
  $^{1}$SenseTime Research\\
  $^{2}$Institute of Automation, CAS; $^{3}$Peng Cheng Lab
%   $^{2}$Qing Yuan Research Institute, Shanghai Jiao Tong University, Shanghai, China\\
}
  
%   $^{2}$Qing Yuan Research Institute, Shanghai Jiao Tong University, Shanghai, China\\
%   \texttt{\{jinguoqiang, yangfan1, sunmingshan, liuyakun1, liwei1,} \\
%   \texttt{baotianpeng, zhaorui, wuliwei\}@sensetime.com} 

% First Author\\
% Institution1\\
% Institution1 address\\
% {\tt\small firstauthor@i1.org}
% For a paper whose authors are all at the same institution,
% omit the following lines up until the closing ``}''.
% Additional authors and addresses can be added with ``\and'',
% just like the second author.
% To save space, use either the email address or home page, not both
% \and
% Second Author\\
% Institution2\\
% First line of institution2 address\\
% {\tt\small secondauthor@i2.org}
% }
\maketitle

% To this end, we propose a novel transformer-based self-supervised learning method for object detection, which is based on the sequence consistency between the online and momentum branches.
%%%%%%%%% ABSTRACT
\begin{abstract}
 
Self-supervised pre-training and transformer-based networks have significantly improved the performance of object detection. 
However, most of the current self-supervised object detection methods are built on convolutional-based architectures.
We believe that the transformers' sequence characteristics should be considered when designing a transformer-based self-supervised method for the object detection task.
To this end, we propose \textbf{SeqCo-DETR}, a novel \textbf{Seq}uence \textbf{Co}nsistency-based self-supervised method for object \textbf{DE}tection with \textbf{TR}ansformers.
SeqCo-DETR defines a simple but effective pretext by minimizes the discrepancy of the output sequences of transformers with different image views as input and leverages bipartite matching to find the most relevant sequence pairs to improve the sequence-level self-supervised representation learning performance.
Furthermore, we provide a complementary mask strategy incorporated with the sequence consistency strategy to extract more representative contextual information about the object for the object detection task.
Our method achieves state-of-the-art results on MS COCO (45.8 AP) and PASCAL VOC (64.1 AP), demonstrating the effectiveness of our approach.
\end{abstract}

%%%%%%%%% BODY TEXT
\section{Introduction}
\label{sec:intro}
Object detection is a prediction-intensive process compared with the image classification task, which needs to locate and classify multiple objects in an image \cite{liu2020deep}. 
Existing deep learning-based object detection frameworks can be divided into one-stage methods \cite{tian2019fcos,law2018cornernet} and two-stage methods \cite{ren2015faster,cai2018cascade}, either of them requires hand-crafted components. 
Recently, transformer-based detection methods show a new object detection paradigm \cite{CarionMSUKZ20,Zhu2021deform}, which is a full end-to-end process without hand-crafted components. 
Compared with convolutional-based architectures, transformer-based architectures define the problem as a sequence-to-sequence process; that is, the transformer converts the input to a sequence and processes the information in the form of a sequence, and the final output is also a sequence \cite{vaswani2017attention}.
%
% This sequence-based processing method has been proven to be significantly effective in both natural language processing tasks \cite{devlin2018bert,brown2020language} and computer vision fields \cite{dosovitskiy2020image,han2020survey}. 
%
The transformer-based architectures do not rely on the inductive bias characteristics of convolutional-based architectures, such as locality and translation invariance, but rely on the global information processing procedure based on attention mechanism \cite{vaswani2017attention,dosovitskiy2020image}. 
However, the abovementioned detection methods require supervised training, which needs massive labeled data with extensive human labor since the labeling cost of object detection tasks is much higher than the image classification tasks.

\begin{figure}[t]
	\centering
	\includegraphics[width=\linewidth]{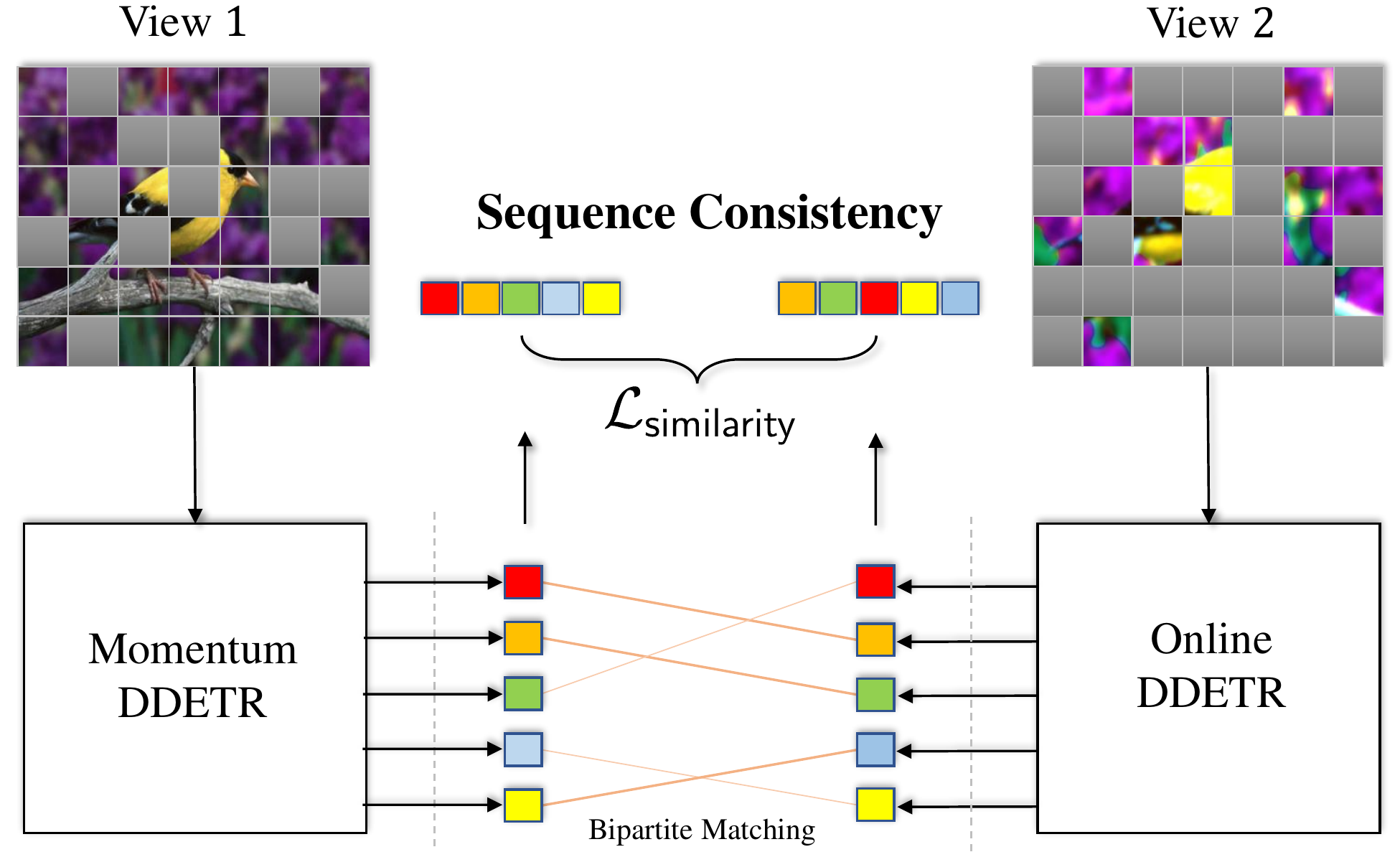}
	\caption{Illustration of the sequence consistency strategy and the complementary mask strategy proposed in our SeqCo-DETR. Note that each output sequence, denoted as colored squares in the figure, contains the object feature and location information.}
	\label{fig:one1}
	 \vspace{-10pt}
\end{figure}

Self-supervised representation learning could leverage unlabeled data efficiently by requiring the model to solve customized pretext tasks on unlabeled data \cite{he2020momentum,chen2021empirical,chen2020simple,CaronMMGBJ20,grill2020bootstrap}. 
By introducing the customized pretext tasks, self-supervised representation learning methods could pre-train a model to make it equipped with the feature representation ability. After the pre-training process, the learned feature representation ability of a model could be beneficial to solve specific downstream tasks.
Most self-supervised methods are designed for image classification tasks, which consider the image as a whole and only use image-level features \cite{abs-2102-04803}. 
However, object detection is a prediction-intensive task that requires predicting the location and category of multiple objects in one image, which needs object-level features. 
Therefore, directly applying these image-level self-supervised methods to objection detection tasks would lead to limited improvement. 
Some recent approaches \cite{abs-2102-04803,abs-2011-13677,yang2021instance,DBLP:journals/corr/abs-2106-11952,roh2021spatially,wang2021dense,wu2021align,xie2021propagate} utilize the inductive bias characteristics of the convolutional neural network (CNN) to achieve object-level self-supervised learning, which is unsuitable for the transformers. 
More recently, some transformer-based pre-training methods have begun to emerge \cite{DaiCLC21, bar2022detreg}. However, they handle the pre-training task in an unsupervised way, using hand-crafted pseudo labels to supervise the pre-training process, which would limit the feature representation ability of the model.

% UP-DETR \cite{DaiCLC21} utilized random patches as the input to predict the location of the patches. Since the location of the patch is known, the proposed pretext task is more like a supervised method with pseudo labels. DETReg \cite{bar2022detreg} relies on mimicking pre-trained model features; thus, the feature representation's ability is limited by the selected pre-trained model, limiting the model's performance. 
% %However, they only address the object-level feature learning while missing the self-supervised learning on the location regression task.

To solve the aforementioned problems, we take advantage of the sequence characteristics of transformers and propose a self-supervised object detection pre-training method (SeqCo-DETR) by maintaining consistency of the sequence from different views of an image, as illustrated in \cref{fig:one1}. 
Each output sequence of the transformer decoder stands for an object prediction, which contains the location and category information of the object. Thus, the proposed pretext task addresses self-supervised learning on both the location and category of objects, which are the two essential tasks of object detection.
Considering that the object prediction of each sequence varies under different image views, we propose to utilize bipartite matching \cite{kuhn1955hungarian} to get the optimal sequence pair to improve the sequence consistency learning process.
Since the object detection task needs to locate the object according to the context information around the object, we propose to add complementary masks on different image views, to help the model to learn more global context information about the object.
The proposed SeqCo-DETR is based on the Deformable DETR (DDETR) \cite{Zhu2021deform} object detection framework; thus, we could pre-train the entire object detection architecture end-to-end, not only the backbone part.

% Moreover, we add complementary masks on different image views,  thus the paired object prediction sequences from different branches would base on totally different local features, forcing the network to pay more attention to the global context information of the object, not just the local features. 

In summary, our main contributions are: 

\begin{itemize}
    \item The proposed SeqCo-DETR is a transformer-based self-supervised learning method for object detection by maintaining the sequence consistency between different image views, which leverages the sequence characteristics of the transformers, and achieves state-of-the-art results on various benchmarks.
	
    \item The complementary mask augmentation strategy is proposed to help the model to extract more representative global context information of the object, which is designed to incorporate with the proposed sequence consistency strategy.
    
	\item The bipartite matching is adopted to get the optimal sequence pairs from the online and momentum branches with different image views, which boosts the performance of the proposed method.
    
% 	The visualization results show that our approach could implicitly realize the position consistency by ensuring the consistency of the output tokens.
\end{itemize}

%
% Since the object detection task needs to locate the object according to the context information, we add complementary masks on different image views, thus the paired object prediction sequences from different branches would base on totally different local features, forcing the network to pay more attention to the global context information of the object, not just the local features. 
%

% Moreover, we add complementary masks on different image views,  thus the paired object prediction sequences from different branches would base on totally different local features, forcing the network to pay more attention to the global context information of the object, not just the local features. 

\section{Related Work}\label{sec:related}
\noindent\textbf{Transformer-based object detection methods.} 
The DEtection TRansformers (DETR) \cite{CarionMSUKZ20}  brings a new paradigm of the object detection task, which is a fully end-to-end method without hand-crafted components. DETR is based on the encoder-decoder transformers, which does not rely on the inductive bias of convolutional-based architectures and defines the object detection problem as a set prediction problem. However, the training speed of the original DETR is considerably slow, and the detection results on small objects are limited. To solve the problems, Deformable DETR \cite{Zhu2021deform} uses multiscale features and proposes a deformable attention mechanism, which significantly accelerates the convergence speed and improves the overall accuracy. Therefore, our method is based on the Deformable DETR framework.

\noindent\textbf{Self-supervised representation learning.} 
Instance discrimination is one of the competitive pretext tasks for self-supervised visual representation learning, which aims to learn such an embedding space in which similar sample pairs stay close to each other while dissimilar ones are far apart \cite{wu2018unsupervised}. MoCo \cite{he2020momentum,chen2020improved, chen2021empirical} improved the training of contrastive methods by storing representations in a momentum structure. SimCLR \cite{chen2020simple} proved that the memory bank can be replaced with large batch size and more image augmentations. SwAV \cite{CaronMMGBJ20} took the features as a set of trainable clustering prototype vectors. BYOL \cite{grill2020bootstrap} utilized the asymmetric architecture together with the stop gradient design to bootstrap the representations by extracting features from different views of the same instance, which could be trained without negative samples. Although these methods have shown promising performance in image classification tasks, they have limited improvement on prediction-intensive tasks such as object detection \cite{abs-2102-04803}.

\noindent\textbf{Self-supervised object detection methods.} 
In order to improve the object detection task via self-supervised learning, several methods have been proposed recently. DetCo \cite{abs-2102-04803} proposed a contrastive loss between local patches and the global image to improve the multi-level feature expression ability for detection tasks. ORL \cite{DBLP:journals/corr/abs-2106-11952} achieved object-level representation based on scene images. SCRL \cite{roh2021spatially}, ReSim \cite{xiao2021region}, DetCon \cite{henaff2021efficient}, ContrastiveCrop \cite{peng2022crafting}, and SoCo \cite{wei2021aligning} proposed to maintain region-level consistency in the related areas by different methods. DenseCL \cite{wang2021dense} and PixPro \cite{xie2021propagate} proposed to utilize the pixel-level to achieve dense contrastive learning. Self-EMD \cite{abs-2011-13677} proposed to utilize the spatial information of CNN features and used Earth Mover’s Distance to match features from different views. InsLoc \cite{yang2021instance} and Align Yourself \cite{wu2021align} pasted cropped images at different locations, then minimized the corresponding features extracted from different views, while it failed to consider the localization task in the object detection. However, these methods rely on the inductive bias characteristics of CNN, which is not suitable for the transformers. 

More recently, UP-DETR \cite{DaiCLC21} designed an unsupervised pre-train pretext based on the transformer architecture, which utilized random patches as the input to predict the location of the patches. Since the location of the patch is known, the proposed pretext task is more like a supervised method with pseudo labels. DETReg \cite{bar2022detreg} proposed to use the Selective Search \cite{uijlings2013selective} to generate region proposals as the location supervision instead of random proposals, whereas using a pre-trained model as the feature supervision. DETReg relies on the mimicking of pre-trained model features, thus the feature representations ability is limited by the pre-trained model, which would limit the model's performance. Therefore, in our approach, we propose to use the self-supervised method to learn the feature representation via maintaining the consistency of sequences of transformers instead of learning the features from a fixed pre-trained model.

\begin{figure*}[!ht]
	\centering
	\includegraphics[width=1\textwidth]{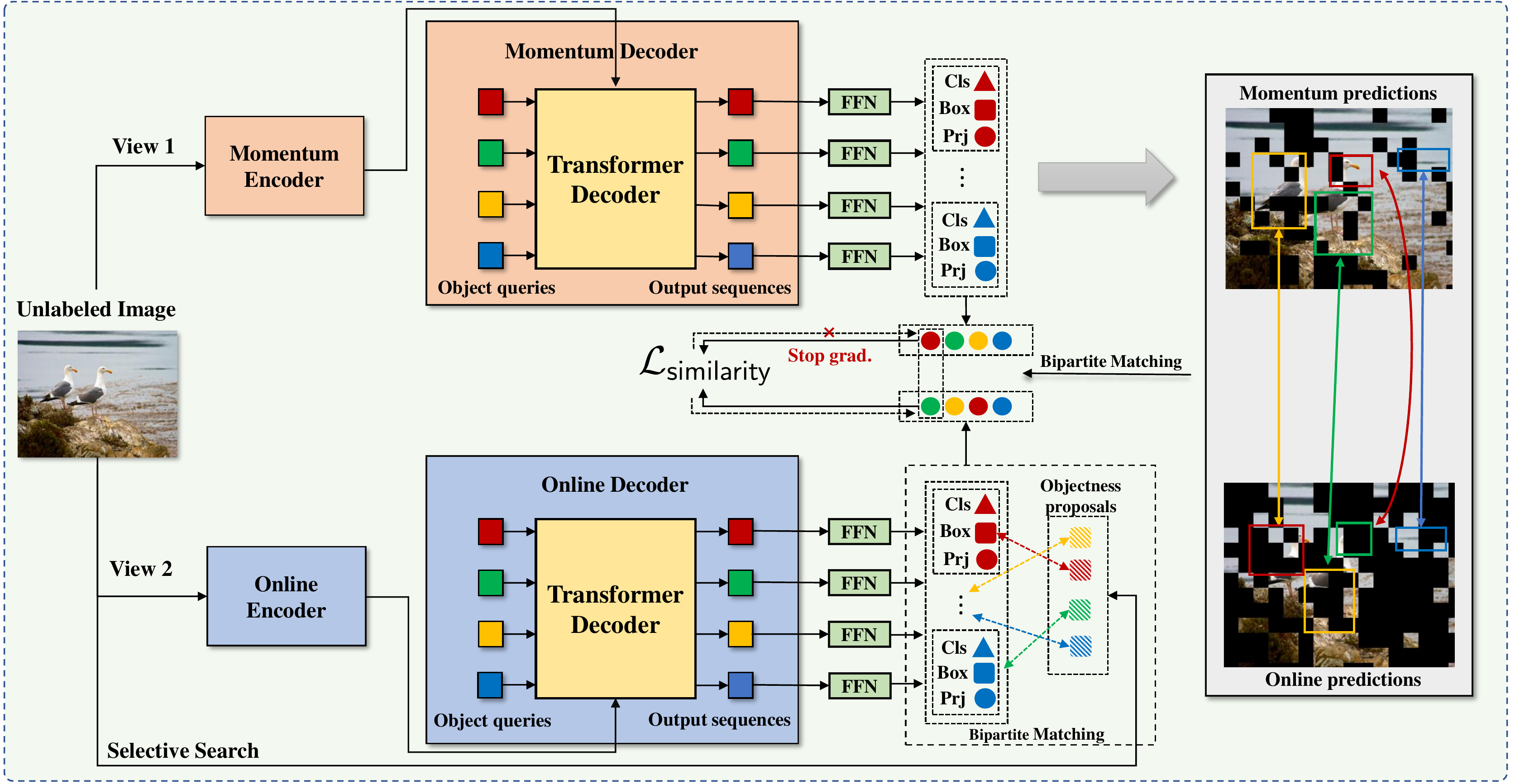}
	\caption{The proposed SeqCo-DETR contains two branches: the online branch and momentum branch. The Online/Momentum Encoder in the figure consists of a CNN backbone and a transformer encoder. The gradient only updates the weight of the online branch. The input of each branch is a different view of the unlabeled image. Notably, we add the complementary image masks to each view that each branch only sees a non-overlapping part of the image. After being processed by the transformers, all object queries become the output sequences. Then the sequences pass three feedforward networks (FFN) to be predicted as the class, bounding box, and feature projection of the object, respectively. There are two bipartite matching processes: one is between the outputs of two branches, the other is between the output of online branches and the ``objectness" proposals provided by the Selective Search. Finally, the similarity loss of each paired sequence feature projections is minimized to achieve sequence-level self-supervised representation learning.}
	\label{fig:mypipeline}
\end{figure*}

\section{Method}
\label{sec:method}

SeqCo-DETR is designed to use the sequence characteristics to achieve the self-supervised pre-training for object detection based on the transformers. The main framework of our SeqCo-DETR is shown in \cref{fig:mypipeline}. To maintain the consistency of sequence, we minimize the discrepancy of the sequences output from the online and momentum branches. The bipartite matching \cite{kuhn1955hungarian} is leveraged to optimize the sequence consistency procedure. These parts will be detailed in \cref{sec:sequence}. Moreover, we design a mask-based image augmentation strategy to force the model to learn the global context information of the object, which will be detailed in \cref{sec:mask}. Then, we describe the way to generate ``objectness" proposals in \cref{sec:rps}. Finally, the whole framework is summarized in \cref{sec:Whole Framework}. Below, we will review these parts in details.

\subsection{Sequence Consistency Strategy}
\label{sec:sequence}

The main idea of SeqCo-DETR is keeping consistency between sequences from differently augmented views of the same image. The main framework is based on the Deformable DETR \cite{Zhu2021deform}. We utilize the momentum design \cite{he2020momentum,xu2021end,liu2021unbiased} to achieve the self-supervised learning, which contains the online branch and the momentum branch. These two branches share the same structure, including CNN backbone, transformer encoder, transformer decoder, and FFN heads, as shown in \cref{fig:mypipeline}. Specifically, we add a projection head \cite{chen2020simple} after the transformer decoder to generate the feature projection for each sequence. The classification and box regression heads are also used during the pre-training process. After pre-training, the projection head will be removed, and the classification head will be reset and modified according to the number of categories in downstream detection tasks. The rest of the weights learned during pre-training will be loaded as the initial weights during fine-turning. Following the momentum design, the weight of the online branch is updated by the gradient, and the weight of the momentum branch is updated by the momentum of the weight of the online branch, according to the formula:
\begin{equation}
	\Theta _{m} \leftarrow \beta \ast \Theta _{m}+(1- \beta)*\Theta _{o},
\end{equation}
where $\Theta _{m}$ and $\Theta _{o}$ represent the parameters of the momentum branch and the online branch, respectively, $ \beta$ represents the updated momentum parameter. Since the momentum branch has a stop gradient design and its parameters are updated by momentum parameters, the two branches would update at different speeds, which can effectively prevent network collapse \cite{grill2020bootstrap}. 

The proposed sequence consistency strategy is simple and straightforward. Since each output sequence stands for an object prediction and each sequence contains the most relevant feature description for each object, we apply the consistency constraint on the sequences that predict the same object. Thus, we could maintain the object-level feature consistency instead of the image-level feature, which is more suitable for object detection tasks. The sequence consistency strategy is formulated as follows:
\begin{equation}
	\label{eq:ssl}
	\mathcal{L}_{\text{ssl}}=\sum_{i=1}^N (\mathcal{L}_{\text{similarity}}(\mathbf{f}_{\text{ffn}} (\mathbf{s}_i),  \mathbf{\widehat{f}}_{\text{ffn}} (\mathbf{\widehat{s}}_{\widehat{\sigma} (i)} ))),
\end{equation}
where $\mathbf{s}$ and $\mathbf{\widehat{s}}$ represent the sequences output by the transformer decoder from the momentum and online branches, respectively; $\mathbf{f}_{\text{ffn}}$ and $\widehat{\mathbf{f}}_{\text{ffn}}$ represent the FFN heads from the momentum and online branches, respectively; $N$ is the number of sequences in one view, and $\widehat{\sigma}$ represents the matching relationship between the two sequences. $\mathcal{L}_{\text{ssl}}$ is the total self-supervised learning loss. $\mathcal{L}_{\text{similarity}}$ is a function that measures the similarity between sequences. We adopt $\mathcal{L}_{2}$ as the $\mathcal{L}_{\text{similarity}}$ loss.

As mentioned above, the image views for the two branches are different, the output sequences from the branches would have diversity too. Thus, we need to match the sequences that have the same object prediction. To this end, we adopt bipartite graph matching \cite{kuhn1955hungarian} to get the optimal sequence pair from the online and momentum branches, which is defined as:
\begin{equation}\small
    \label{eq:match_cost}
	\begin{aligned}
		\mathcal{L}_{\text{match}}(\mathbf{y},\mathbf{\widehat{y}}_{\sigma})=&\sum_{i=1}^{N}[-\lambda _{cm}\log\mathbf{\widehat{p}}_{{\sigma} (i)}(c_{i})+\\
		&\mathbbm{1}_{\left \{ c_{i}\neq \varnothing  \right \}}(\lambda _{bm}\mathcal{L}_{\text{box}}(\mathbf{b}_{i},\mathbf{\widehat{b}}_{{{\sigma} (i)}}))],
	\end{aligned}
\end{equation}
\begin{equation}
	\widehat{\sigma} =\underset{\sigma \epsilon \sum _{N}}{\arg\min}\sum_{i}^{N}\mathcal{L}_{\text{match}}(\mathbf{y}_{i},\mathbf{\widehat{y}}_{\sigma (i)}),
\end{equation}
where $N$ is the number of sequences in one view; $\mathcal{L}_{\text{match}}$ denotes the Hungarian matching loss; $\mathbf{y}$ and $\widehat{\mathbf{y}}$ are the predicted sequences from momentum and online branches, respectively; $\mathbf{b}$ and $c$ are the location prediction and category prediction, respectively; $\widehat{\mathbf{p}}_{\widehat{\sigma} (i)}(c_{i})$ is the probability of class $c_{i}$; $\widehat{\sigma}$ denotes the final optimal assignment; $\varnothing$ denotes the empty set; $\lambda _{cm}$ and $\lambda _{bm}$ are the corresponding weights, which are 2.0 and 5.0, respectively. 

The bipartite graph matching takes consider of location and category of the predicted object from the sequence. In our approach, the categories of the predicted object during the pre-training stage are only foreground or background, as detailed in \cref{sec:rps}. Thus, we mainly focus the location of the predicted object, that is, the sequences used for calculating the $\mathcal{L}_{\text{ssl}}$ are from the objects that are predicted at the same location. Specifically, the two image views share the same location augmentation parameters, and there are only content differences between the two views, such as color, mask, and blur, which reduces the difficulty of outputs sequences matching. The comparison results for the different sequence utilization strategies are summarized in \cref{tab:Sequence utilization method}.

\subsection{Mask strategy}
\label{sec:mask}
\begin{figure}
     \centering
     \begin{subfigure}[b]{0.44\linewidth}
         \centering
         \includegraphics[width=\linewidth]{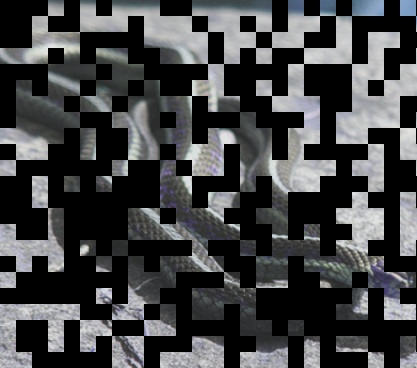} 
         \caption{Mask 50\%}
         \label{fig:mask50}
     \end{subfigure}
    %  \hfill
     \begin{subfigure}[b]{0.44\linewidth}
         \centering
         \includegraphics[width=\linewidth]{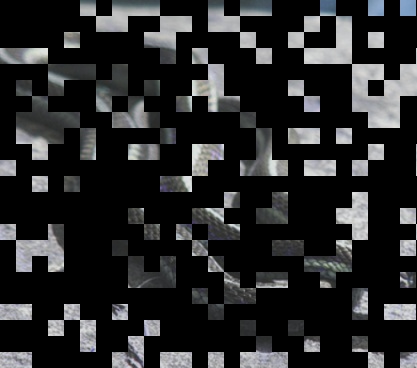} 
         \caption{Mask 70\%}
         \label{fig:mask70}
     \end{subfigure}
     \\
     \begin{subfigure}[b]{\linewidth}
         \centering
         \includegraphics[width=0.44\linewidth]{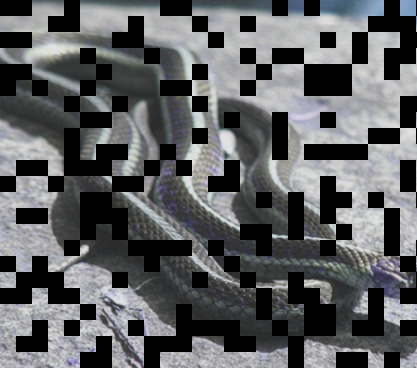}
         \includegraphics[width=0.44\linewidth]{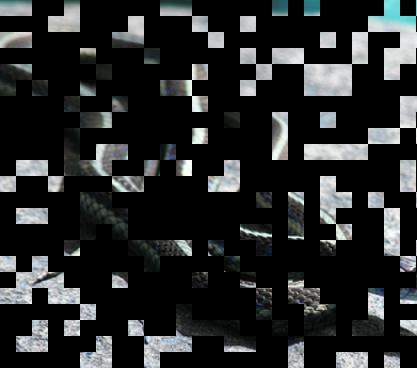}
         \caption{Complementary mask of 30\% and 70\%}
         \label{fig:maskr70}
     \end{subfigure}
        \caption{Examples of different proportions of random masked images. }
        \label{fig:masks}
        \vspace{-15pt}
\end{figure}

The object localization task is critical in object detection, which requires not only the features from the object but also the context information around the object. Therefore, it is necessary to design a strategy to improve the model's global context information extraction capability in the self-supervised object detection mission. Recently, mask-based image augmentation has been proved an efficient way to extract global context information, especially combined with transformers \cite{he2021masked,el2021large,zhou2021ibot}. Due to there being usually redundant information in the local area of an image, adding a mask to the image could force the network to use more faraway context information to extract features, which could improve network global feature extraction capabilities. To this end, we design a mask-based image augmentation incorporated with the proposed sequence consistency strategy to improve the self-supervised training. The examples of random masked images are shown in \cref{fig:masks}. Since the gradient updates only the online branch, there are usually more strong image augmentations used in the online branch \cite{xu2021end,liu2021unbiased}. Following this setting, we design different mask strategies. One straightforward way is to add the mask to only the online branch view, since the online branch needs to explore more while the momentum branch predicts more stable results. 

To further force the network to exploit the contextual information of the object, inspired by \cite{el2021large}, we design a complementary mask strategy that adds complementary masks on both branches. Specifically, with the complementary masks, each image view will not have overlapped area with each other, as shown in \cref{fig:maskr70}; thus the predictions from each branch will not depend on the same local patches, but with the global context information. By minimizing the sequence consistency loss, the network will be guided to extract and combine the global context information to predict the objects, which would be helpful in the object detection task. The comparison results for the different mask strategies are summarized in \cref{tab:masks}.

\subsection{Objectness Proposals}
\label{sec:rps}

As noted previously, each output sequence of the transformer stands for an object prediction, and our self-supervised loss is built upon the features that are projected from the output sequence, as denoted in \cref{eq:ssl}. Thus, the features extracted from the image are critical in our approaches. In order to get more rich semantic features rather than some background features from an image, the transformer needs to predict more ``objectness" proposals, which contain foreground objects as many as possible. Selective Search \cite{uijlings2013selective} considers the local similarity of image textures to generate candidate proposals and those proposals are with higher ``objectness" recall than the random patch in an image, which is an unsupervised way to provide ``objectness" proposals. Selective Search could only generate foreground proposals, thus, there will be only two categories, i.e., foreground and background. Therefore, we leverage the generated initial foreground proposals to train the neural network to have the ability to predict more ``objectness" proposals. The parameters of the Selective Search are the same as in DETReg \cite{bar2022detreg}. The comparison results for the region proposal strategy are summarized  in \cref{tab:different pretrain}.

\subsection{The Whole Framework}
\label{sec:Whole Framework}

The total loss of SeqCo-DETR consists of the region proposals loss $\mathcal{L}_{RPS}$ and the self-supervised learning loss on sequences $\mathcal{L}_{SSL}$, denoted as:
\begin{equation}
	\label{eq:total_loss}
	\begin{aligned}
		\mathcal{L}_{\mathrm{total}}\mathbf{(y,\widehat{y})}=&\mathcal{L}_{RPS}+\mathcal{L}_{SSL} \\
		=&\sum_{i=1}^{N}[\lambda_{f}\mathcal{L}_{\text{focal}}(c_{i},\widehat{\mathbf{p}}_{\widehat{\sigma} (i)}) + \\
		&\mathbbm{1}_{\left \{ c_{i}\neq \varnothing  \right \}}(\lambda_{b}\mathcal{L}_{\text{box}}(\mathbf{b}_{i},\widehat{\mathbf{b}}_{{\widehat{\sigma} (i)}})]+\\
		&\sum_{i=1}^{N}[\lambda_{e}\mathcal{L}_{\text{ssl}}(\mathbf{z}_{i}, \widehat{\mathbf{z}}_{\widehat{\sigma} (i)}))],
	\end{aligned}
\end{equation}
where $\mathcal{L}_{\text{focal}}$ is the focal loss of classification, $\mathcal{L}_{\text{box}}$ is the location loss of box, $\mathcal{L}_{\text{ssl}}$ is the self-supervised learning loss of sequence, and $\mathbf{z}_{i}$ stands for the output of $\mathbf{f}_{\text{ffn}}$. The $\mathcal{L}_{RPS}$ consists of $\mathcal{L}_{\text{focal}}$ and $\mathcal{L}_{\text{box}}$, which is the same as in DETReg \cite{bar2022detreg}, supervised by the proposals from Selective Search. $\lambda_{f}$, $\lambda_{b}$, and $\lambda_{e}$ are the weights of those three losses, which are set to 2.0, 5.0, and 10.0, respectively. In summary, the algorithm flow of the proposed method can be summarized in \cref{alg:baseline}.

\definecolor{codeblue}{rgb}{0.25,0.5,0.5}
\definecolor{keyword}{rgb}{0.8, 0.25, 0.5}
\newcommand{\listingsttfamily}{\fontfamily{pcr}\small}
\lstset{
	backgroundcolor=\color{white},
	basicstyle=\fontsize{8pt}{8pt}\listingsttfamily,
	columns=fullflexible,
	breaklines=False,
	commentstyle=\fontsize{8pt}{8pt}\color{codeblue},
	keywordstyle=\fontsize{8pt}{8pt}\color{keyword},
}
\begin{figure}
	\centering
		\begin{minipage}{\linewidth}
\begin{algorithm}[H]
    \setstretch{0.8}
	\caption{Pseudo code of SeqCo-DETR in a PyTorch-like style.}
	\label{alg:baseline}
	\begin{adjustbox}{width=\linewidth}
	\begin{lstlisting}[language=python,tabsize=4,showtabs]
# model_momentum: momentum backbone 
# + encoder + decoder + head
# model_online: online backbone 
# + encoder + decoder + head 
# mse: mean squared error, i.e., L2 loss
# augment: image augmentation, 
# more details are in Sec. 4.1
# matcher: bipartite matching algorithm
# rps: region proposals from Selective Search
# criterion: classification and regression loss

for param in model_momentum.parameters():
    param.requires_grad = False
    
def similarity_loss(e1, e2, ids_ssl):
    loss = mse(e1-e2[ids_ssl).sum().mean()
    return loss

for x in dataloader:  # load a batch x with B samples
    x1, x2 = weak_augment(x), strong_augment(x)
    
    with torch.no_grad():
        cls_m, box_m, prj_m = model_momentum(x1)
        
    cls_o, box_o, prj_o = model_online(x2)
    
    idx_rps = matcher((cls_o, box_o), rps)
    loss_rps = criterion((cls_o, box_o), rps, idx_rps)  
    
    idx_ssl = matcher((cls_m, box_m), (cls_o, box_o))
    loss_ssl = similarity_loss(prj_m, prj_o, idx_ssl)
    
    loss = loss_rps + loss_ssl
    loss.backward()
    
    model_online.update()
    model_momentum.update()    
\end{lstlisting}
	\end{adjustbox}
\end{algorithm}
	\end{minipage}
 	\vspace{-5pt}
\end{figure}

\section{Experiments}

\subsection{Implementation Details}

\noindent\textbf{Datasets.} 
Our experiments include the pre-training stage and the fine-tuning stage. First, we pre-train models on the unlabeled dataset. Then, we load the pre-trained weight and fine-tune the network on the downstream object detection tasks following the standard procedure. For the pre-training stage, in order to be consistent with the results of DETReg \cite{bar2022detreg}, we use ImageNet (IN1K) \cite{deng2009imagenet} and ImageNet100 (IN100) \cite{DengDSLL009} as the main pre-training datasets. In particular, ImageNet100 is a subset of ImageNet, which only contains 100 classes, a total of about 125k images. The split of classes is the same as in DETReg. For COCO results, the pre-training dataset is IN1K; for VOC results, the pre-training dataset is IN100; For the few-shot task, the pre-training dataset is IN1K. The ablation studies are all based on the IN100. The Selective Search \cite{uijlings2013selective}, based on OpenCV \cite{bradski2000opencv}, is used to generate the initial foreground proposals, which only contains two categories. We only use the top 30 proposals in one image, which is the same as DETReg. In the fine-tuning stage, we evaluate our method on MS COCO \cite{LinMBHPRDZ14} and PASCAL VOC \cite{EveringhamGWWZ10}. In particular, we fine-tune the model on COCO \texttt{train2017} and evaluate it on COCO \texttt{val2017}. As for VOC, we fine-tune on VOC \texttt{trainval07+12}, and then evaluate on \texttt{test07}. In addition, we also pre-train our model on multi-objects dataset on COCO and COCO+ in the ablation studies, where COCO stands for \texttt{train2017} without ground truth labels, COCO+ denotes the COCO \texttt{train2017} plus the COCO \texttt{unlabeled} dataset. And we also use Selective Search to generate initial foreground proposals for the two unlabeled multi-objects datasets.

\noindent\textbf{Training details.} 
In the pre-training stage, we add an additional FFN head for projecting sequence feature. The FFN is composed of 2 hidden layers, and the number of hidden layer neurons is 256. In addition, since all the proposals generated by Selective Search are foreground boxes, the classification head only have 2 categories, i.e., foreground and background. The backbone of the proposed method is ResNet-50 \cite{he2016deep}, which is initialized by SwAV \cite{CaronMMGBJ20}, which is the same as in DETReg \cite{bar2022detreg}. More training details are provided in the Supplementary file.

% Following DETReg, the number of epochs in pre-training is 50, the batch size is 24, and the initial learning rate is $2 \cdot 10^{-4}$, which is decayed after 40 epochs by a factor of 10. For IN1K, the pre-training epoch is 5. The parameters in the fine-tuning stage are also the same with DETReg. On the COCO, the epoch is 50, the batch size is 4, and the initial learning rate is $2 \cdot 10^{-4}$, which is decayed after 40 epochs by a factor of 10. On the VOC, the epoch is 100, the batch size is 4, and the initial learning rate is $2 \cdot 10^{-4}$, which is decayed after 70 epochs by a factor of 10. For the few-shot object detection task, we follow the standard protocol \cite{wang2020frustratingly} as used in DETReg. Experiments are carried out on 8 * NVIDIA V100 GPUs.

\noindent\textbf{Image augmentations.} 
\label{sec:aug}
As mentioned in \cref{sec:mask}, we design a complementary mask strategy to incorporate with the sequence consistency strategy. Thus, the image augmentations for the two branches are different, where weak image augmentations are used in the momentum branch while strong augmentations are used in the online branch. The complementary mask proportions for the online and momentum branches are 70\% and 30\%, respectively, with the patch size of 16, which are all chosen by experiments. To ensure the two views have consistency in the location parameters, the image $view~\#2$ is partially built upon the $view~\#1$. Following \cite{deng2021unbiased}, we first generate a base image view from the input unlabeled image, using random flips, random resize, and random resized-crop. The base image view is the same as in DETReg \cite{bar2022detreg}. Then, for the momentum branch, we add the corresponding mask upon the base image view to generate $view~\#1$. As for the $view~\#2$, we add more augmentations upon the base image view, including color jitter, random grayscale, random blur, and the corresponding mask. Thus, there is no location coordinates difference between the two views. 

\subsection{Main Results}
\label{sec:Main Results}
We compare the performance of SeqCo-DETR and other methods on  object detection and few-shot object detection tasks. The comparison approaches are DETReg \cite{bar2022detreg}, UP-DETR \cite{DaiCLC21} based on Deformable DETR, Deformable DETR \cite{Zhu2021deform} with different pretrained weights, and the common baseline Faster R-CNN \cite{ren2015faster}.

\begin{table*}[htb]
	\centering
	%\resizebox{\linewidth}{!}{
		\begin{tabular}{@{}l|lll||lllll@{}}
			\toprule
			                            & \multicolumn{3}{c||}{COCO \texttt{val2017}}&\multicolumn{3}{c}{VOC \texttt{test07}} \\ 	
			Model                            & AP   & AP${}_{50}$ & AP${}_{75}$ & AP  & AP${}_{50}$  & AP${}_{75}$ \\ \midrule
			Faster R-CNN \cite{lin2017feature} ~                      & 42.0 & 62.1 & 45.5 & 56.1 & 82.6 & 62.7\\
			%DETR\cite{CarionMSUKZ20}                             & 42.0 & 62.4 & 44.2 & 20.5 &  45.8 & 61.1 \\
	       % UP-DETR\cite{DaiCLC21}                            & 42.8 & 63.0 & 45.3 & 20.8 & 47.1 & \textbf{61.7} \\		
			 
	        Deformable DETR (Supervised CNN) \cite{Zhu2021deform}     & 43.8 & 62.6 & 47.7 & 59.5 & 82.6 & 65.6 \\	\midrule	

			Deformable DETR (SimCLR CNN)${}^{\dag}$        &   41.5  &  59.8     &  45.4   &  57.3   &  80.0   &  63.6    \\		        
			Deformable DETR (BYOL CNN)${}^{\dag}$        &   44.7  &  63.8     &  48.8   &  59.9   &  82.7   &  66.7    \\	
			Deformable DETR (MoCo CNN)${}^{\dag}$        &   43.1  &  61.6    &  46.9   &  59.6   &  81.8   &  66.0    \\		
			Deformable DETR (SwAV CNN)${}^{\dag}$        &   45.0  & 63.8     &  49.2     &   61.0 & 83.0 & 68.1    \\	
			
			UP-DETR (Deformable DETR)   ${}^{\ddag}$                      & 44.7    & 63.7    &  48.6    & 61.8   & 83.4 & 69.6   \\

			DETReg w/o feature embedding ${}^{\dag}$  & 45.2 & 63.7 & 49.5 & 63.0 & 83.5 & 70.2                           \\
			DETReg \cite{bar2022detreg}                             & 45.5     & 64.1     &  49.9  & 63.5 & 83.3 & 70.3    \\ \midrule
            SeqCo-DETR                & \textbf{45.8} & \textbf{64.2} & \textbf{50.0} & \textbf{64.1} & \textbf{83.8} & \textbf{71.6}  \\ 
            % SeqCo-DETR (ours)                & \textbf{45.7} & \textbf{64.2} & \textbf{49.9} & 26.6 & \textbf{49.4}&60.0 \\ 			
			\bottomrule
		\end{tabular}
		%}
	\caption{
		Comparison results on the object detection datasets MS COCO \texttt{val2017} and PASCAL VOC \texttt{test07}.
		\dag: We run the method on our codebase.	
		\ddag: Results are provided by DETReg\cite{bar2022detreg}.
	}
	\label{tab:coco result}		
	\vspace{-5pt}
\end{table*}

\noindent\textbf{Results on object detection task with MS COCO and PASCAL VOC Datasets.}
The experiment results are shown in Table \ref{tab:coco result}. The above part of the table is the widely referenced baselines in object detection tasks, which are list here for convenience. The middle part is our baseline, and the bottom part is our approach. 
Since the training process of DETReg is supervised by the pre-trained SwAV model \cite{CaronMMGBJ20}, the pre-training process can be regarded as learning the fixed features from SwAV. On the contrary, our method proposes to use the self-supervised ways to learn the features, which could help the network learn more discriminative features during pre-training. 
When the feature embedding learning part of DETReg is removed in the experiment, the final results are 45.2 on COCO and 63.0 on VOC, which could be regarded as a baseline. 
For the final result, our method achieves 45.8 and surpasses DETReg by 0.3 points on COCO and 0.6 points on VOC, proving that our learning-based method has better performance than the method that is supervised by manually defined pseudo-labels. 
Compared to the methods that only the backbone part is pre-trained, i.e., Deformable DETR (Supervised CNN) and Deformable DETR with different self-supervised pre-trains such as SimCLR \cite{chen2020simple}, BYOL \cite{grill2020bootstrap}, MoCo v2 \cite{chen2020improved}, and SwAV \cite{CaronMMGBJ20}, our method could pre-train the entire object detection framework, and surpasses Deformable DETR (SwAV CNN) by 0.8 points on COCO and 3.1 points on VOC. Thus, our approach is more suitable for the transformer-based object detection task.

\noindent\textbf{Results on few-shot object detection task.}
Following the standard procedure in the few-shot setting \cite{wang2020frustratingly, bar2022detreg}, we split the COCO dataset into 60 base classes and 20 novel classes. The 60 base classes contains the full data, while there are only $k\in \{10, 30\}$ in each novel class. We first fine-tune the model on the base classes using the default parameters on COCO. Then, we fine-tune it on the base and novel classes, the same as in DETReg. Specifically, for $k=10$, we fine-tune 30 epochs with learning rate of $2 \cdot 10^{-5}$, for $k=30$, we fine-tune 50 epochs with learning rate of $4 \cdot 10^{-5}$. The results are reported on the novel classes. As shown in \cref{tab:few-shot}, the final results are similar to those on COCO and VOC. Our SeqCo-DETR outperforms DETReg by 2.0 and 0.9 points in $k=10$ and $k=30$ settings, respectively, proving the effectiveness of the proposed sequence-based self-supervised representation learning method. 

\begin{table}
% \vspace{-15pt}
\centering
\resizebox{\linewidth}{!}{
\begin{tabular}{l|cccc}
\toprule
\multirow{2}{*}{Model} & \multicolumn{2}{c}{~Novel AP} & \multicolumn{2}{c}{Novel AP${}_{75}$} \\ 
                        & 10            & 30           & 10             & 30            \\ \midrule
FRCN+ft-full \cite{wang2020frustratingly} & 9.2           & 12.5         & 9.2            & 12.0          \\
% TFA \cite{wang2020frustratingly}                  & 10.0          & 13.7         & 9.3            & 13.4          \\
Deformable DETR (Supervised CNN) \ddag                 & 23.3          & 28.4         & 25.4           & 31.7          \\  \midrule

% Deformable DETR (SimCLR CNN)${}^{\dag}$        &   21.1  &  20.3     &  23.2   &  22.0    \\		        
% Deformable DETR (BYOL CNN)${}^{\dag}$        &     &       &     &       \\	
% Deformable DETR (MoCo CNN)${}^{\dag}$        &  18.9  &  18.7    &  20.4   &  20.4       \\		
% % Deformable DETR (SwAV CNN)${}^{\dag}$        &   24.9  & 25.8     &  27.3    &   27.9     \\	
% Deformable DETR (SwAV CNN)${}^{\dag}$        &   24.9  & 25.8     &  27.3    &   27.9     \\	
UP-DETR (Deformable DETR)   ${}^{\dag}$                &   23.9      &   27.1     &   26.3        &  29.4       \\ 
DETReg w/o feature embedding ${}^{\dag}$  & 24.2 & 26.1 & 26.5 & 28.2 \\

DETReg \cite{bar2022detreg}                & 25.0          & 30.0         & 27.6           & \textbf{33.7}          \\  \midrule
SeqCo-DETR      &  \textbf{27.0}           &    \textbf{30.9}        &     \textbf{29.7}           &   33.4           \\ \bottomrule
\end{tabular}
}
\caption{Comparison results on few-shot detection task on COCO, evaluated on the novel classes. \dag: We run the method on our codebase. \ddag: Results are provided by DETReg \cite{bar2022detreg}.}
\label{tab:few-shot}
 \vspace{-5pt}
\end{table}

\subsection{Ablation Study}

% The mask proportions used for the main experiments are chosen by several experiments, which are the best values, more details are in the Supplementary file. 

\noindent\textbf{Mask strategy.}
As noted before, we design a complementary mask strategy that adds complementary masks on different image views to force the model to learn to extract the global context information during the self-supervised pre-training stage. As shown in \cref{fig:masks}, we conduct extensive experiments about the mask strategy. A straightforward way is to add the mask to the online branch, since the online branch is updated by the gradient. When the mask is only added to the online branch, with the mask proportion of 50\%, denoted as $\text{Mask}_{online @50}$, there is no improvement in our method. Meanwhile, when adding the same mask to DETReg, the result drops from 45.4 to 45.0. The mask strategy may interfere with the training process that is only supervised by the pseudo labels. When the complementary masks are used in online and momentum branches, with mask proportion of 70\% and 30\%, respectively, denoted as $\text{Mask}_{online @70} + \text{Mask}_{\neg({online @70})}$, we get the best value of 45.8. To verify whether the complementary characteristic is the most critical design in the mask design, we add the independent random masks for the two branches with the same mask proportion of 70\% and 30\%, respectively, denoted as $\text{Mask}_{online @70} + \text{Mask}_{momentum @30}$. The corresponding result is 45.6. We also try the same proportion masks for the two branches, with the proportion of 50\%, denoted as $\text{Mask}_{online @50} + \text{Mask}_{momentum @50}$. The result is only 45.4. These two experiments prove that the complementary mask is a more effective way compared to only adding random masks to the two branches. Moreover, based on the sequence consistency strategy, combining the complementary mask will get better performance. More experiments on mask parameter selection can be found in the Supplementary file.

% The mask proportions used for the main experiments are chosen by several experiments, which are the best values, more details are in the Supplementary file. 

% Though, the mask strategy improves the results of UP-DETR from 44.7 to 45.0, the final result still lower than ours. 
% This proves that the mask strategy is more suitable in our self-supervised approach. 

% The mask strategy may interfere with the training process that is only supervised by the pseudo labels. 

% When the image mask is only used in the online branch, with mask proportion of 50\%, denoted as $\text{Mask}_{online @50}$, the result of our method improves from 45.2 to 45.6. Meanwhile, when we add the same mask to the DETReg, the result drops from 45.4 to 45.0. 

% Though, the mask strategy improves the results of UP-DETR from 44.7 to 45.0, the final result still lower than ours. 
% This proves that the mask strategy is more suitable in our self-supervised approach. 

% The mask strategy may interfere with the training process that is only supervised by the pseudo labels. 

\begin{table}
% \vspace{-15pt}
\centering
\resizebox{\linewidth}{!}{
\begin{tabular}{@{}c|l|l@{}}
\toprule
Model                       & Mask strategy                          & AP   \\ \midrule
\multirow{2}{*}{DETReg}     & w/o Mask (baseline) \cite{bar2022detreg}                           & 45.4 \\
                            & w/ $\text{Mask}_{50}$  ${}^{\dag}$                               & 45.0 \\ \midrule
% \multirow{2}{*}{ \begin{tabular}[c]{@{}c@{}}UP-DETR\\  (Deformable DETR) \end{tabular}   }     & w/o Mask (baseline) ${}^{\dag}$                            & 44.7 \\
%                             & w/ $\text{Mask}_{50}$  ${}^{\dag}$                              & 45.0 \\ \midrule                            
\multirow{5}{*}{SeqCo-DETR} & w/o Mask                     & 45.6 \\
&  $\text{Mask}_{online @50}$                     & 45.6 \\
                            % & Feature mask      & 45.6 \\
                            & $\text{Mask}_{online @50} + \text{Mask}_{momentum @50}$       & 45.4 \\
                            & $\text{Mask}_{online @70} + \text{Mask}_{momentum @30}$ & 45.6 \\
                            & $\text{Mask}_{online @70} + \text{Mask}_{\neg({online @70})}$         & \textbf{45.8} \\ 
                            % & Complementary mask + feature mask      & 45.3 \\                             
                            \bottomrule
\end{tabular}
}
\caption{Comparison of mask strategies, evaluated on MS COCO \texttt{val2017}. \dag: We run the method on our codebase.}
\label{tab:masks}
 \vspace{-5pt}
\end{table}

% We conduct several experiments to verify the performance on the multi-object datasets. 

\noindent\textbf{Pre-training datasets and region proposal strategy.}
We conduct several experiments to compare the performance with different types of pre-training datasets. \texttt{Rnd bbox} stands for the random proposals, COCO GT stands for the region proposals come from the COCO ground truth. For COCO and COCO+, we use the images without ground truth, and use Selective Search to generate initial proposals. Specifically, COCO is a multi-object dataset, while ImageNet is a single-object dataset. As shown in \cref{tab:different pretrain}, our method achieves better results against DETReg on both single-object and multi-object datasets, which proves that self-supervised learning based has better generality over different types of datasets. Furthermore, we compare the influence of different region proposal strategies. When using the less ``objectness" proposals, i.e., random proposals, both method drops a lot. When using the ground truth as the region proposal, DETReg improves from 45.1 to 45.6, whereas ours improves from 45.6 to 45.8. This proves that a method that is fully dependent on the hand-craft pseudo labels would be easily affected by the quality of the pseudo labels, whereas the learning-based method has less affected by the pseudo labels. The quality of the region proposals generated by Selective Search is sufficient for the proposed method to learn useful information.

\begin{table}
% \vspace{-15pt}
\centering
\resizebox{\linewidth}{!}{
\begin{tabular}{c|ccccc}
\toprule
Method     & IN100          &   IN100 (Rnd bbox)        & COCO & COCO+ & COCO GT \\ \midrule
DETReg ${}^{\dag}$     & 45.4 &44.1 & 45.1 & 45.1  & 45.6                         \\
SeqCo-DETR & \textbf{45.8} &  \textbf{44.3}      & \textbf{45.6} & \textbf{45.6}  & \textbf{45.8}                         \\ \bottomrule
\end{tabular}
}
\caption{Comparison of pre-training datasets and region proposal strategie, evaluated on MS COCO \texttt{val2017}. \dag: We run the method on our codebase.}
\label{tab:different pretrain}
 \vspace{-10pt}
\end{table}

% The proposed sequence consistency strategy aims to maintain the output sequence consistency between the online and momentum branches to achieve the self-supervised pre-training. One of the most naive strategies is the one-by-one match, because the sequence output by transformers has sequential characteristics. However, different branches have different input image views and slightly different network parameters. Even if the output sequences are in the same order, their predicted results are still different. Therefore, the bipartite matching is adopted to match the sequences from different branches which predict the same object.

\noindent\textbf{Sequence utilization methods.}
As mentioned before, considering the importance of the sequences that are output by transformers, we design a sequence consistency strategy that maintains the output sequence consistency between the online and momentum branches to achieve the self-supervised pre-training. One of the most naive strategies is the one-by-one match, because the sequence output by transformers has sequential characteristics. However, different branches have different input image views and slightly different network parameters. Even if the output sequences are in the same order, their predicted results are still different. Therefore, the bipartite matching is adopted to match the sequences from different branches which predict the same object. As shown in \cref{tab:Sequence utilization method}, bipartite matching could improve results from 45.6 to 45.8, compared to the one-by-one matching, which proves the effectiveness of bipartite matching method. Meanwhile, in order to achieve a more sufficient supervision of the sequence, we try to use the outputs of classification head $f_{\text{cls}}$, regression head  $f_{\text{box}}$, and projection head $f_{\text{prj}}$ in \cref{eq:ssl} at the same time. It can be seen from the table that the fusion of multiple heads decreases the final results in both matching settings. Maybe the self-supervision on the projection head is enough; redundancy supervision on three heads would cause a performance drop. 

\begin{table}
    % \vspace{-15pt}
	\centering
	\resizebox{\linewidth}{!}{
	\begin{tabular}{l|ccc|c}
		\toprule
		Model & \begin{tabular}[c]{@{}c@{}}One-by-one\\ matching\end{tabular}    &\begin{tabular}[c]{@{}c@{}}Bipartite\\ matching\end{tabular}  & Multi-feature & AP \\
		\midrule
		\multirow{4}{*}{SeqCo-DETR} & \checkmark & & &  45.6 \\
		& \checkmark  && \checkmark &45.3 \\
		&   &\checkmark& \checkmark  & 45.5 \\
		&  &\checkmark &   & \textbf{45.8}  \\
		\bottomrule
	\end{tabular}
}
	\caption{Comparison of sequence utilization strategies, evaluated on MS COCO \texttt{val2017}.}
	\label{tab:Sequence utilization method}
% \vspace{-10pt}
\end{table}

\noindent\textbf{Self-supervised loss strategy.}
As shown in \cref{tab:contrastive loss result}, we compare the effects of different self-supervised losses. Predictor + BYOL loss follows the BYOL \cite{grill2020bootstrap} architecture, which adds an extra MLP layer on the online branch to make the two branches asymmetrical and the inputs of online and momentum branches are also exchanged to create symmetry loss. MoCo loss comes from MoCo v3 \cite{chen2021empirical} symmetric version. Surprisingly, L1 and L2 loss can achieve satisfactory results, thus, the L2 loss is adopted in the experiments.

\begin{table}
    % \vspace{-15pt}
	\centering
	\begin{tabular}{@{}l|l|l@{}}
		\toprule
		Model                            & Self-supervised loss strategy & AP   \\ \midrule
		\multirow{4}{*}{SeqCo-DETR} & L1 loss                   & 45.6 \\
		& L2 loss                   & \textbf{45.8} \\
		& MoCo loss                   & 45.4 \\
		& Predictor + BYOL loss     & 45.2 \\ \bottomrule
	\end{tabular}
	\caption{Comparison of self-supervised loss strategies on MS COCO \texttt{val2017}. }
	\label{tab:contrastive loss result}	
	\vspace{-10pt}
\end{table}

\section{Conclusion}
In this paper, we present SeqCo-DETR, a novel transformer-based self-supervised learning method for object detection, which takes advantage of the sequence characteristics of the transformers to achieve self-supervised learning of detection by maintaining the consistency of the sequence under different image views. Furthermore, we propose a complementary mask strategy to incorporate sequence consistency to extract more global context information to improve the object detection task. The bipartite matching is leveraged to get the optimal sequence pair to improve the efficiency of sequence-level self-supervision. Extensive experiments on various downstream detection tasks and on both single-object and multi-object datasets prove the effectiveness of the SeqCo-DETR.
% Extensive experiments prove the effectiveness of the SeqCo-DETR.
% These two experiments prove that the complementary mask is a more effective way compared to only adding random masks to the two branches. Moreover, based on the sequence consistency strategy, combining the complementary mask will get better performance. More experiments on mask parameter selection can be found in the Supplementary file.
% This proves that a method that is fully dependent on the hand-craft pseudo labels would be easily affected by the quality of the pseudo labels, whereas the learning-based method has less affected by the pseudo labels. The quality of the region proposals generated by Selective Search is sufficient for the proposed method to learn useful information.

%%%%%%%%% REFERENCES
{\small
\bibliographystyle{ieee_fullname}
\bibliography{main}
}

\newpage

\appendix

\setcounter{table}{0}
\renewcommand{\thetable}{A\arabic{table}}

\setcounter{figure}{0}
\renewcommand{\thefigure}{A\arabic{figure}}

% \begin{appendix}
\section{Appendix}
\label{sec:appendix}
\subsection{Overview}
\label{sec:overview}

We organize the supplementary material as follows.
The implementation details are given in \cref{sec:detail}. More results of the ablation study are presented in \cref{sec:more ablation}, especially about the proposed mask strategy. 
Then, we visualize the matched proposal pairs between the two branches in \cref{sec:vis}.
Finally, we discuss the limitations and future work of our approach in \cref{sec:limit}.

\subsection{Implementation Details}
\label{sec:detail}

Our approach contains the pre-training and fine-turning stages. Following DETReg \cite{bar2022detreg}, in the pre-training stage, for the ImageNet100 (IN100) \cite{DengDSLL009} dataset, the number of epochs is 50, the batch size is 24, and the initial learning rate is $2 \cdot 10^{-4}$, which is decayed after 40 epochs by a factor of 10. For the ImageNet (IN1K) \cite{deng2009imagenet} dataset, the pre-training epoch is 5. The parameters in the fine-tuning stage are also the same with DETReg. On Ms COCO \cite{LinMBHPRDZ14}, the number of epoch is 50, the batch size is 4, and the initial learning rate is $2 \cdot 10^{-4}$, which is decayed after 40 epochs by a factor of 10. On PASCAL VOC \cite{EveringhamGWWZ10}, the number of epoch is 100, the batch size is 4, and the initial learning rate is $2 \cdot 10^{-4}$, which is decayed after 70 epochs by a factor of 10. For the few-shot object detection task, we follow the standard protocol \cite{wang2020frustratingly} as used in DETReg. The following ablation studies are all pre-trained on IN100 and fine-tuned on COCO. Experiments are carried out on 8 * NVIDIA V100 GPUs.

\subsection{More results on the mask strategy}
\label{sec:more ablation}
As mentioned in the main paper, we propose a complementary mask strategy for the self-supervised learning method for object detection. In order to determine the best combinations and optimal parameters of the mask strategy, we conduct multiple ablation experiments, which will be detailed below.

\begin{figure}[ht]
    % \vspace{-5pt}
     \centering
     \begin{subfigure}[b]{0.155\textwidth}
         \centering
         \includegraphics[width=\textwidth]{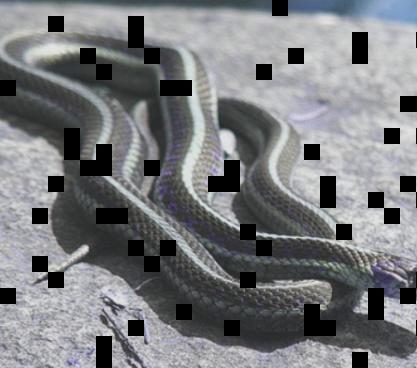} 
         \caption{ 10\%}
         \label{fig:mask5010}
     \end{subfigure}
    %  \hfill
     \begin{subfigure}[b]{0.155\textwidth}
         \centering
         \includegraphics[width=\textwidth]{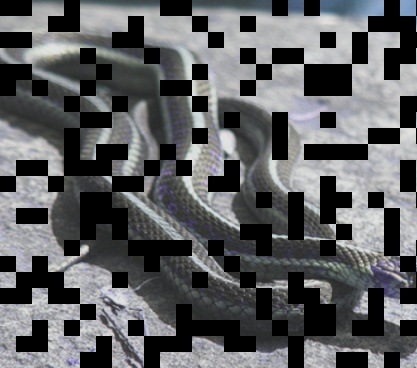} 
         \caption{ 30\%}
         \label{fig:mask7030}
     \end{subfigure}
    %  \hfill
     \begin{subfigure}[b]{0.155\textwidth}
         \centering
         \includegraphics[width=\textwidth]{figures/50.png} 
         \caption{ 50\%}
         \label{fig:mask5050}
     \end{subfigure}
    %  \hfill
     \begin{subfigure}[b]{0.155\textwidth}
         \centering
         \includegraphics[width=\textwidth]{figures/70.png} 
         \caption{ 70\%}
         \label{fig:mask7070}
     \end{subfigure}     
    %  \hfill
     \begin{subfigure}[b]{0.155\textwidth}
         \centering
         \includegraphics[width=\textwidth]{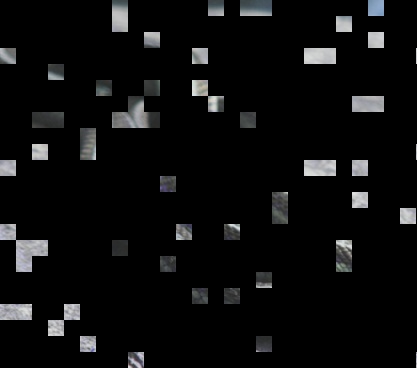} 
         \caption{ 90\%}
         \label{fig:mask5090}
     \end{subfigure}
    %  \hfill
     \begin{subfigure}[b]{0.155\textwidth}
         \centering
         \includegraphics[width=\textwidth]{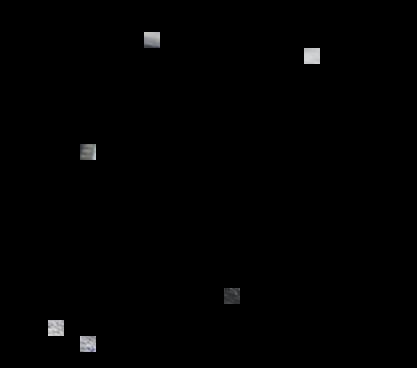} 
         \caption{ 99\%}
         \label{fig:mask7099}
     \end{subfigure}    
     \caption{Illustration of different proportions of masked images.}
     \label{fig:masks_all}
     \vspace{-5pt}
\end{figure}

\subsubsection{Single mask strategy for the online branch}

In this section, we conduct several experiments to find the optimal proportion of the single image mask. As shown in \cref{fig:masks_all}, we demonstrate the images after being masked with different proportions. Specifically, we only add the mask to the online branch, and the patch size of the mask is 16. Experiments on the patch size are given in \cref{sec:patch_size_mask}. 

The results are listed in \cref{tab:mask proportion}, where \texttt{Rnd(30,80)} stands for the mask proportions are randomly sampled from 30\% to 80\%. From the table, the results of ours stay the same at first and then decreases as the mask proportion increases. We also try to use the random proportion of mask during pre-training, and the \texttt{Rnd(30,80)} achieves the best result. As a counterpart, the performance of our baseline DETReg \cite{bar2022detreg} decreases when adding masks to the input image. When there is no mask added to the image, i.e., the mask proportion is 0, the default result is 45.4, as reported in the paper. When there are different proportions of masks added to DETReg, the performance drops, and the results prove that the mask strategy is not suitable for DETReg. Because DETReg is fully dependent on the hand-craft pseudo labels, which will be easily influenced. But the mask strategy is suitable for our SeqCo-DETR, because our approach employs the self-supervised method to learn features. The combination of the image mask and the sequence consistency strategies could extract more representative contextual information about the object.

\begin{table}[]
% \vspace{-5pt}
\centering
\resizebox{\linewidth}{!}{
\begin{tabular}{@{}l|cccccccc}
\toprule
\begin{tabular}[c]{@{}c@{}}Mask\\  proportion(\%)\end{tabular}  & 0    & 30    & 50    & 70    & 90  & 99  & Rnd(30,80) & Rnd(0,99) \\ \midrule
DETReg      & 45.4 & - & 45.0 & - & - &- &45.3 &-        \\ 
SeqCo-DETR      &\textbf{45.6}  &\textbf{45.6} & \textbf{45.6} & 45.5  & 45.3 & 45.0 &\textbf{45.7} &45.3        \\ \bottomrule
\end{tabular}
}
\caption{Comparison of different proportions of single masks added to the online branch, evaluated on MS COCO \texttt{val2017}.}
\label{tab:mask proportion}
% \vspace{-15pt}
\end{table}

\subsubsection{Different mask strategies for the two branches}
% In this section, we study which kind of mask to add. 

\noindent\textbf{Two branches with complementary masks}
We add the complementary masks to both branches, as in the main paper. The complementary mask means the mask for each branch is strictly binary reversed. Thus, each branch will have a non-overlapped image view.

\begin{table}
% \vspace{-15pt}
\centering
\resizebox{\linewidth}{!}{
\begin{tabular}{@{}c|l|l@{}}
\toprule
Model                       & Mask strategy                          & AP   \\ \midrule
\multirow{5}{*}{SeqCo-DETR} & $\text{Mask}_{online @30} + \text{Mask}_{\neg({online @30})}$                      & 45.3 \\
&  $\text{Mask}_{online @50} + \text{Mask}_{\neg({online @50})}$                    & 45.1 \\
                     
                            &$\text{Mask}_{online @70} + \text{Mask}_{\neg({online @70})}$      & \textbf{45.8} \\
                            & $\text{Mask}_{online @75} + \text{Mask}_{\neg({online @75})}$ & 45.5 \\
                            &$\text{Mask}_{online @80} + \text{Mask}_{\neg({online @80})}$ & 45.2 \\
                            % & $\text{Mask}_{online @70} + \text{Mask}_{momentum @30}$ & \textbf{45.6} \\  
                            \bottomrule
\end{tabular}
}
\caption{Comparison of different proportions of the complementary mask added to both branches, evaluated on MS COCO \texttt{val2017}. }
\label{tab:mask_com}
\end{table}

The results are listed in \cref{tab:mask_com}, where $\text{Mask}_{online @30} + \text{Mask}_{\neg({online @30})}$ stands for the mask proportion for the online branch is 30\% while for the momentum branch is 70\%, where the mask for the momentum branch is the complementary mask of the online branch. From the table, the best proportion for the online branch is 70\%, correspondingly, for the momentum branch is 30\%, which is the complementary mask of the online branch.

\noindent\textbf{Two branches with random masks} 
We also try to add random masks to both branches, including each branch having the same proportions and having different proportions of masks. Especially, the mask for each branch is independently sampled.  

\begin{table}
% \vspace{-15pt}
\centering
\resizebox{\linewidth}{!}{
\begin{tabular}{@{}c|l|l@{}}
\toprule
Model                       & Mask strategy                          & AP   \\ \midrule
\multirow{8}{*}{SeqCo-DETR} & $\text{Mask}_{online @30} + \text{Mask}_{momentum @30}$                     & 45.2 \\
&  $\text{Mask}_{online @50} + \text{Mask}_{momentum @50}$                     & 45.4 \\
                     
                            & $\text{Mask}_{online @70} + \text{Mask}_{momentum @70}$       & 45.4 \\\cmidrule{2-3}
                            & $\text{Mask}_{online @20} + \text{Mask}_{momentum @80}$ & 45.1 \\
                            & $\text{Mask}_{online @30} + \text{Mask}_{momentum @70}$ & 45.1 \\
                            & $\text{Mask}_{online @70} + \text{Mask}_{momentum @30}$ & \textbf{45.6} \\  
                            & $\text{Mask}_{online @75} + \text{Mask}_{momentum @25}$ & 45.5 \\   
                            & $\text{Mask}_{online @80} + \text{Mask}_{momentum @20}$ & \textbf{45.6} \\                               
                            \bottomrule
\end{tabular}
}
\caption{Comparison of different proportions of random mask added to both branches, evaluated on MS COCO \texttt{val2017}. }
\label{tab:mask_two}
% \vspace{-15pt}
\end{table}

As listed in \cref{tab:mask_two}, when the masks are independently sampled for both branches with different proportions, the best parameter is  $\text{Mask}_{online @70} + \text{Mask}_{momentum @30}$, the corresponding result is 45.6, which is lower than the complementary mask, as 45.8. This proves the complementary mask strategy is better. We also try to add the same proportion masks for the two branches, but the results are lower than those with different proportions or the complementary proportions of masks. These experiments prove that the complementary mask strategy is more suitable for our method.
% Besides, the experiment results also demonstrate that when there are more proportions of mask added to the momentum branch, the results are lower.

\subsubsection{The patch size of the mask}
\label{sec:patch_size_mask}
\begin{figure}[]
     \centering
     \begin{subfigure}[b]{0.155\textwidth}
         \centering
         \includegraphics[width=\textwidth]{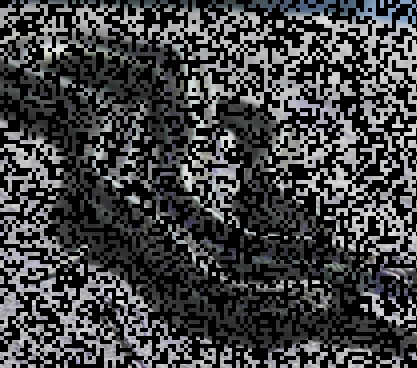} 
         \caption{ 4}
         \label{fig:p4}
     \end{subfigure}
    %  \hfill
     \begin{subfigure}[b]{0.155\textwidth}
         \centering
         \includegraphics[width=\textwidth]{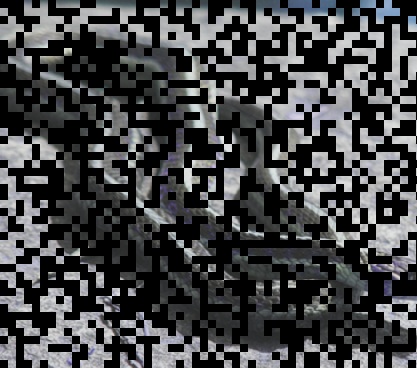} 
         \caption{ 8}
         \label{fig:p8}
     \end{subfigure}
    %  \hfill
     \begin{subfigure}[b]{0.155\textwidth}
         \centering
         \includegraphics[width=\textwidth]{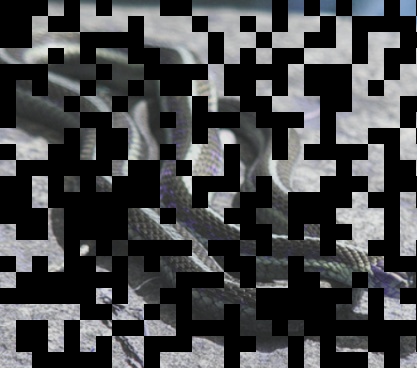} 
         \caption{16}
         \label{fig:p16}
     \end{subfigure}
    %  \hfill
     \begin{subfigure}[b]{0.155\textwidth}
         \centering
         \includegraphics[width=\textwidth]{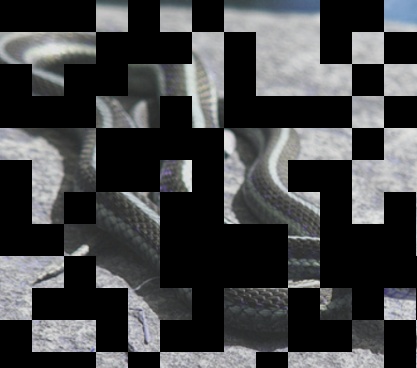} 
         \caption{ 32}
         \label{fig:p32}
     \end{subfigure}     
    %  \hfill
     \begin{subfigure}[b]{0.155\textwidth}
         \centering
         \includegraphics[width=\textwidth]{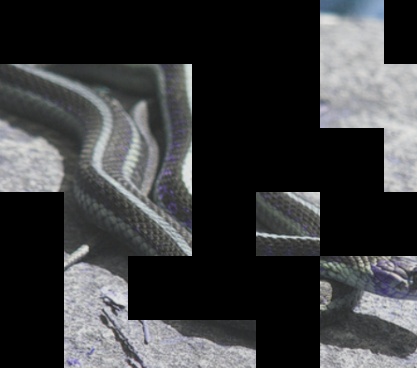} 
         \caption{64}
         \label{fig:p64}
     \end{subfigure}
    %  \hfill
     \begin{subfigure}[b]{0.155\textwidth}
         \centering
         \includegraphics[width=\textwidth]{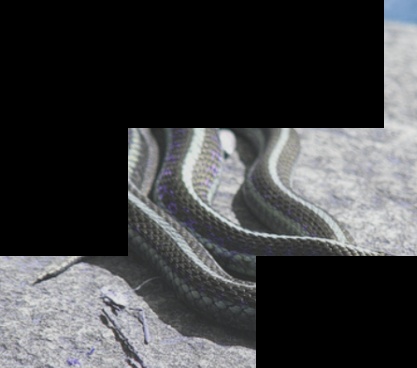} 
         \caption{ 128}
         \label{fig:p128}
     \end{subfigure}    
     \caption{Illustration of different patch sizes of masked images.}
     \label{fig:patch_all}
\end{figure}

In this section, we investigate the patch size of the mask. The masked images are shown in \cref{fig:patch_all}. The image size is randomly resized between 320 and 480 during pre-training. The proportion of image mask is 50\% in the following experiments. As listed in \cref{tab:all_patches}, when the patch sizes are 8 or 16, the final results are the best. The final result seems less affected by the patch size, since the variance of the final result is small when patch sizes vary in a wide range. To choose the best value between patch sizes 8 and 16, we conduct other experiments. When the patch size is 8 and with the mask proportions of \texttt{Rnd(30,80)}, the result is 45.5, which is lower than 45.7 with patch size 16. When the patch size is 8 and with $\text{Mask}_{online @70} + \text{Mask}_{\neg({online @70})}$, the result is 45.5, which is lower than 45.8 with 16. Based on the above experiments, we choose the patch size of 16 as the optimum parameter.

\begin{table}[]
% \vspace{-5pt}
\centering
\resizebox{\linewidth}{!}{
\begin{tabular}{@{}l|cccccccc}
\toprule
Patch size & 4    & 8    & 16   & 32   & 64   & 128  & Rnd(8,32)& Rnd(4,64)\\ \midrule
SeqCo-DETR & 45.3 & \textbf{45.6} & \textbf{45.6} & 45.4 & 45.4 & 45.4&45.3 &45.5\\ \bottomrule
\end{tabular}
}
\caption{Comparison of different patch sizes of mask added to the online branch, evaluated on MS COCO \texttt{val2017}.}
\label{tab:all_patches}
\end{table}

\begin{figure*}[ht]
	\centering
	\includegraphics[width=1\textwidth]{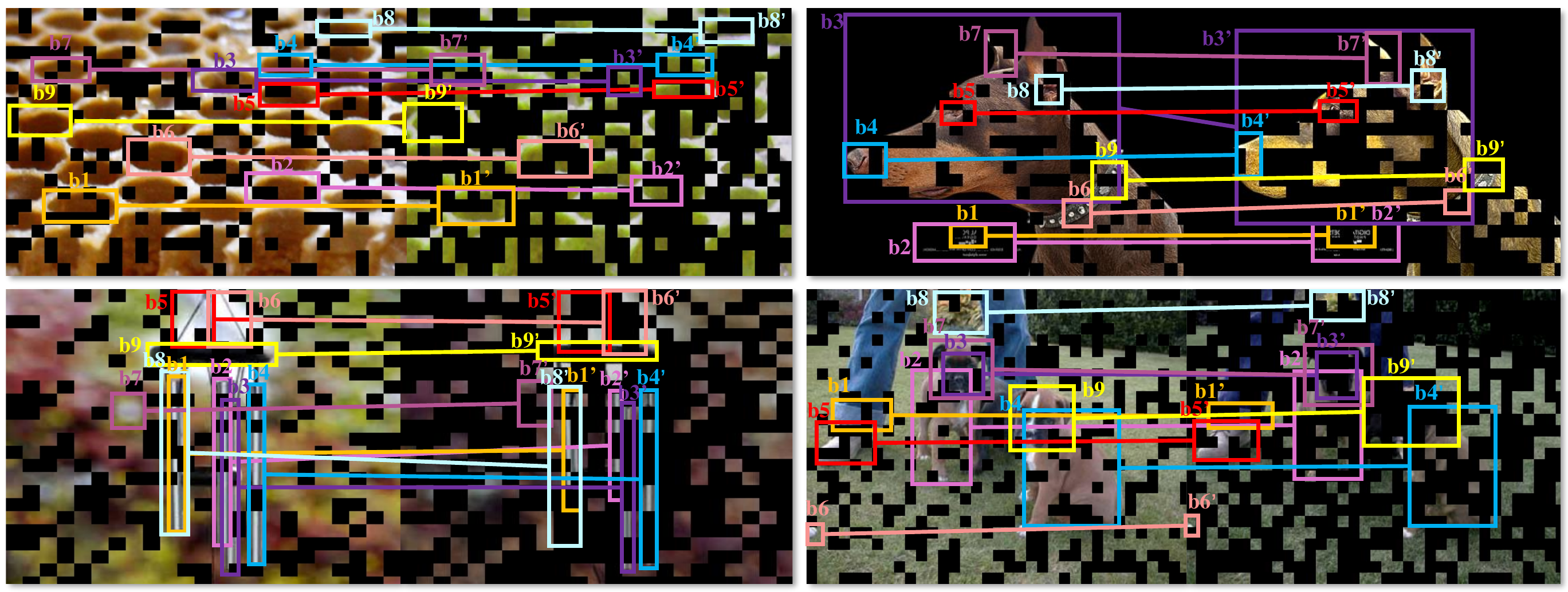}
	\caption{Visualization of paired proposals from different branches. The left is from the online branch, and the right is from the momentum branch.}
 	\label{fig:vis_match}
\end{figure*}

\subsubsection{The feature mask}
We try to add the mask to the features that are involved in calculating the self-supervised loss. Specifically, the feature mask is added to the output sequences from the projection head, and the feature mask is implemented by Dropout \cite{srivastava2014dropout}. As listed in \cref{tab:feature_mask}, the best proportion of feature mask value is 20\%, with the result of 45.6. 

\begin{table}
\centering
\resizebox{\linewidth}{!}{
\begin{tabular}{@{}l|cccccc@{}}
\toprule
Feature mask proportion(\%) & 10   & 20   & 30   & 50   & 70   \\ \midrule
SeqCo-DETR                  & 45.5 & \textbf{45.6} & 45.2 & 45.3 & 45.3 \\ \bottomrule
\end{tabular}
}
\caption{Comparison of different proportions of feature mask added to the online branch, evaluated on MS COCO \texttt{val2017}.}
\label{tab:feature_mask}
\end{table}

Furthermore, we try to combine the feature mask and the image mask. We combine each best value of the image mask and the feature mask. As listed in \cref{tab:f_i}, where $\text{Mask}_{online @50} + \text{Mask}_{feature@20}$ stands for the image mask with 50\% proportion and feature mask with 20\% proportion, $\text{Mask}_{complementary @70-30} + \text{Mask}_{feature @1}$ stands for the complementary mask with 70\% proportion for online branch and the corresponding 30\% proportion for the momentum branch and feature mask with 1\% proportion. However, the combination results show there are huge performance drops when combining the two strategies. For example, when using  $\text{Mask}_{complementary @70-30}$, i.e., $\text{Mask}_{complementary @70-30} + \text{Mask}_{feature @0}$, and $\text{Mask}_{feature@20}$ alone, the result is 45.8 and 45.6, respectively, but the combination result is only 45.2.  Moreover, the results show that the final results become higher when there are fewer proportions of the feature mask. So it is not suitable to combine the image mask and the feature mask. One possible reason is that the combination would increase the difficulty of feature representation learning, which leads to performance drops. Thus, we only adopt the image mask strategy in the final version.

\begin{table}
% \vspace{-15pt}
\centering
\resizebox{\linewidth}{!}{
\begin{tabular}{@{}c|l|l@{}}
\toprule
Model                       & Mask strategy                          & AP   \\ \midrule
\multirow{6}{*}{SeqCo-DETR} & $\text{Mask}_{online @50} + \text{Mask}_{feature@20}$                     & 45.2 \\
&  $\text{Mask}_{online @Rnd(30,80)} + \text{Mask}_{feature @20}$                     & 45.2 \\
                            & $\text{Mask}_{complementary @70-30} + \text{Mask}_{feature @50}$ & 45.1\\            
                            & $\text{Mask}_{complementary @70-30} + \text{Mask}_{feature @20}$    & 45.2  \\
                            & $\text{Mask}_{complementary @70-30} + \text{Mask}_{feature @10}$  & 45.3 \\
                            & $\text{Mask}_{complementary @70-30} + \text{Mask}_{feature @1}$ & 45.4 \\
                             & $\text{Mask}_{complementary @70-30} + \text{Mask}_{feature @0}$ & \textbf{45.8} \\                      
                            \bottomrule
\end{tabular}
% \vspace{-5pt}
}
\caption{Comparison of different combinations of mask strategies, evaluated on MS COCO \texttt{val2017}.}
\label{tab:f_i}
\end{table}

\subsection{Visualization}
\label{sec:vis}
As is shown in \cref{fig:vis_match}, we visualize the matched proposal pairs between the two branches. Only part of the proposals are visualized for clarity and brevity, and the matching indexes are derived from the bipartite matching. From the figure, the matching result is reliable between the two branches. The predicted bounding boxes from the two branches are on the same object, since the matching is mainly based on the location of the predicted bounding box. The mask added to the image has less affection for the matching, whereas the boundary of the predicted bounding box is more likely to be in the unmasked area, as shown in the right part of each image pair. Notably, we are not using the feature just inside a bounding box, but the feature from the sequence. The predicted bounding box is projected from the sequence, since each sequence stands for an object prediction. And the sequence has not only the feature of the object, but also the global context information thanks to the characteristic of the transformer. The visualization proves the proposed method could achieve constraints on output sequences that predict the same object.

\subsection{Limitation and discussion}
\label{sec:limit}
In the bipartite matching for the sequence consistency training, the two views need to share the same crop window location, which limits our method with complex image augmentations. We have not achieved a fully self-supervised detection pre-training, but still rely on the help of the Selective Search. We believe future works will break these limitations.

\end{document}